\documentclass[10pt,twocolumn,letterpaper]{article}

\usepackage[]{cvpr}       
\usepackage[ruled,linesnumbered]{algorithm2e}
\usepackage{multicol}
\usepackage{multirow}
\usepackage{bm}
\usepackage{amsmath} 
\usepackage{bbding}
\usepackage{pifont}
\usepackage{colortbl}
\usepackage[accsupp]{axessibility}
\usepackage[table]{xcolor}
\usepackage{newfloat}
\usepackage{listings}
\definecolor{cvprblue}{rgb}{0.21,0.49,0.74}
\usepackage[pagebackref,breaklinks,colorlinks,allcolors=cvprblue]{hyperref}

\def\paperID{6292} 
\def\confName{CVPR}
\def\confYear{2026}

\title{Mitigating Error Amplification in Fast Adversarial Training}

\author{
  Mengnan Zhao$^{1}$, Lihe Zhang$^{2}$\thanks{corresponding author.}, Bo Wang$^{2}$, 
  Tianhang Zheng$^{3*}$, Hong Zhong$^{1}$, Geyong Min$^{4}$\\
  $^{1}$Anhui University, Anhui, China \quad
  $^{2}$Dalian University of Technology, Liaoning, China \\
  $^{3}$Zhejiang University, Zhejiang, China \quad
  $^{4}$University of Exeter, UK
}

\begin{document}

\maketitle
\begin{abstract}
Fast Adversarial Training (FAT) has proven effective in enhancing model robustness by encouraging networks to learn perturbation-invariant representations.
However, FAT often suffers from catastrophic overfitting (CO), where the model overfits to the training attack and fails to generalize to unseen ones. Moreover, robustness-oriented optimization typically leads to notable performance degradation on clean inputs, and such degradation becomes increasingly severe as the perturbation budget grows.
In this work, we conduct a comprehensive analysis of how guidance strength affects model performance by modulating perturbation and supervision levels across distinct confidence groups.
The findings reveal that low-confidence samples are the primary contributors to CO and the robustness–accuracy trade-off. Building on this insight, we propose a Distribution-aware Dynamic Guidance (DDG) strategy that dynamically adjusts both the perturbation budget and supervision signal. Specifically, DDG scales the perturbation magnitude according to the sample confidence at the ground-truth class, thereby guiding samples toward consistent decision boundaries while mitigating the influence of learning spurious correlations. Simultaneously, it dynamically adjusts the supervision signal based on the prediction state of each sample, preventing overemphasis on incorrect signals. 
To alleviate potential gradient instability arising from dynamic guidance, we further design a weighted regularization constraint.
Extensive experiments on standard benchmarks demonstrate that DDG effectively alleviates both CO and the robustness–accuracy trade-off.
\end{abstract}

\section{Introduction}
\label{sec:intro}
Improving the adversarial robustness of deep neural networks remains a fundamental challenge in trustworthy machine learning\cite{yu2025mtl,yu2025robust,yu2025backdoor,shao2026baldro,zhai2026maximizing}. Among existing defense strategies, adversarial training (AT) has emerged as one of the most effective approaches for enhancing model robustness against adversarial perturbations~\cite{yang2024structure,casper2024defending,zhang2024defensive,yue2024revisiting,fang2024enhancing,tang2024effective}. 
However, standard AT typically incurs substantial computational overhead, as it requires iterative generation of adversarial examples.
To mitigate this issue, Fast Adversarial Training (FAT)~\cite{pan2024adversarial,kim2021understanding,li2022subspace,huang2023fast,park2021reliably,yang2024fast} accelerates training by employing single-step attacks such as FGSM-RS~\cite{Wong2020}.

While computationally efficient, FAT methods often suffer from catastrophic overfitting (CO), where model robustness collapses dramatically after only a few training epochs. Prior studies have attributed this phenomenon to factors such as gradient misalignment~\cite{Andriushchenko2020} and feature pathway divergence~\cite{zhao2024catastrophic}.
In addition, both FAT and standard AT~\cite{wang2019improving, jia2022adversarial} encounter a robustness–accuracy trade-off, where enhancing model robustness typically degrades clean accuracy. To address this, recent methods have introduced various techniques such as prior-guided adversarial initialization~\cite{jia2024improving} and label relaxation~\cite{tong2024taxonomy}. However, the underlying mechanisms driving this trade-off, as well as its relationship to CO, remain insufficiently understood.

In this work, we begin by conducting a systematic analysis of how varying guidance strengths affect FAT performance.
Our analysis reveals that low-confidence and misclassified samples are the primary contributors to both CO and the robustness–accuracy trade-off. Specifically, injecting strong adversarial perturbations into these samples exacerbates CO, as the model tends to leverage perturbation-specific artifacts rather than intrinsic semantic cues—analogous to learning spurious, class-dependent backdoor features. Furthermore, we observe that mitigating erroneous guidance on low-confidence samples simultaneously improves both clean accuracy and adversarial robustness. These insights suggest that enforcing uniform guidance across all samples is inherently suboptimal; instead, the guidance should be  dynamically modulated according to the confidence and prediction state of each input.

Motivated by these insights, we propose a Distribution-aware Dynamic Guidance (DDG) strategy. DDG dynamically adjusts perturbation strengths via adaptive budget allocation, which guides samples to reside near similar decision boundaries while mitigating the risk of learning spurious correlations. In addition, DDG modulates supervision signals according to each sample’s prediction state, preventing excessive emphasis on incorrectly predicted classes. To further stabilize optimization, we introduce a weighted regularization term that alleviates potential gradient non-smoothness induced by dynamic guidance.

Our main contributions are summarized as follows:
{\bf (1)} 
We conduct a systematic examination revealing that applying uniform guidance across samples is suboptimal, and that the guidance should adapt to prediction confidence and correctness.
{\bf (2)} Building on this insight, we propose a distribution-aware dynamic guidance (DDG) strategy that dynamically adjusts the perturbation budget and supervision signal according to each sample’s confidence and prediction state.
Furthermore, we introduce a weighted regularization term to stabilize gradients under dynamic guidance.
{\bf (3)} Extensive experiments on standard benchmarks demonstrate that DDG achieves superior robustness and clean accuracy compared to state-of-the-art FAT methods.

\section{Related Work}
\label{sec:formatting}

Deep neural networks have raised security concerns due to their susceptibility to adversarial attacks~\cite{Kurakin2017,dong2018boosting,cao2022advdo,gu2022segpgd,zhong2022shadows,yu2025towards}, prompting growing interest in AT techniques~\cite{mo2022adversarial,jia2022adversarial,wang2024revisiting}. Given the training data $(x, y)\sim D_\text{train}$, Madry et al.~\cite{madry2017towards} formulate AT as a min-max optimization problem, 
\begin{equation}\label{eq1}
\min_\theta \mathbb{E}_{(x, y)\sim D_\text{train}} \left[\max_{\delta\in[-\xi, \xi]} \mathcal{L}\left(f(x+\delta), y\right) \right],
\end{equation}
where $f(\cdot)$ denotes the model parameterized by $\theta$, $\xi$ is the maximum perturbation budget, and $\mathcal{L}(\cdot)$ typically represents the cross-entropy loss.
In contrast, regularization-based methods~\cite{sriramanan2021towards,zhang2019theoretically} align the predictions for clean $x$ and adversarial examples $x+\delta$, 
\begin{equation}\label{eq2}
\min_\theta \mathbb{E}_{(x, y)\sim D_\text{train}}\left[ \mathcal{L}\left(f(x), y\right) + \|f(x+\delta) - f(x)\|_2 \right],
\end{equation}
where $\|\cdot\|_2$ denotes the $\ell_2$ norm function.

Unlike standard AT that utilizes multi-step attacks, FAT employs single-step attacks (\textit{e.g.}, FGSM)~\cite{goodfellow2014explaining,cheng2021fast,shafahi2019adversarial} for improving training efficiency. However, FAT may face the CO issue~\cite{rice2020overfitting,de2022make}. To mitigate CO, prior works have explored techniques including random initialization~\cite{Wong2020}, gradient alignment~\cite{Andriushchenko2020}, regularization-based defenses~\cite{niu2022fast,sriramanan2021towards}, smoothed convergence~\cite{zhao2023fast}, and feature activation consistency~\cite{zhao2024catastrophic}. Beyond CO, both FAT and standard AT exhibit a significant trade-off between adversarial robustness and clean classification accuracy~\cite{jia2022boosting}. To mitigate this issue, researchers have introduced several strategies, such as prior-guided adversarial initialization~\cite{jia2024improving}, adaptive step-size allocation based on gradient norms~\cite{huang2023fast}, feature-space regularization~\cite{jia2024revisiting}, and label relaxation~\cite{tong2024taxonomy}.

Despite recent advances, existing methods still suffer from a significant trade-off between clean accuracy and adversarial robustness, while the intrinsic relationship between CO and this trade-off remains unclear. This work 
reveals that low-confidence and misclassified samples are the primary drivers to both CO and the robustness–accuracy trade-off. Hence, we propose a distribution-aware dynamic guidance that dynamically assigns perturbation budgets and supervision signals per sample, replacing the uniform treatment used in prior works.

\section{Proposed Method}
\begin{figure}[t]
    \centering
        \includegraphics[width=0.47\textwidth]{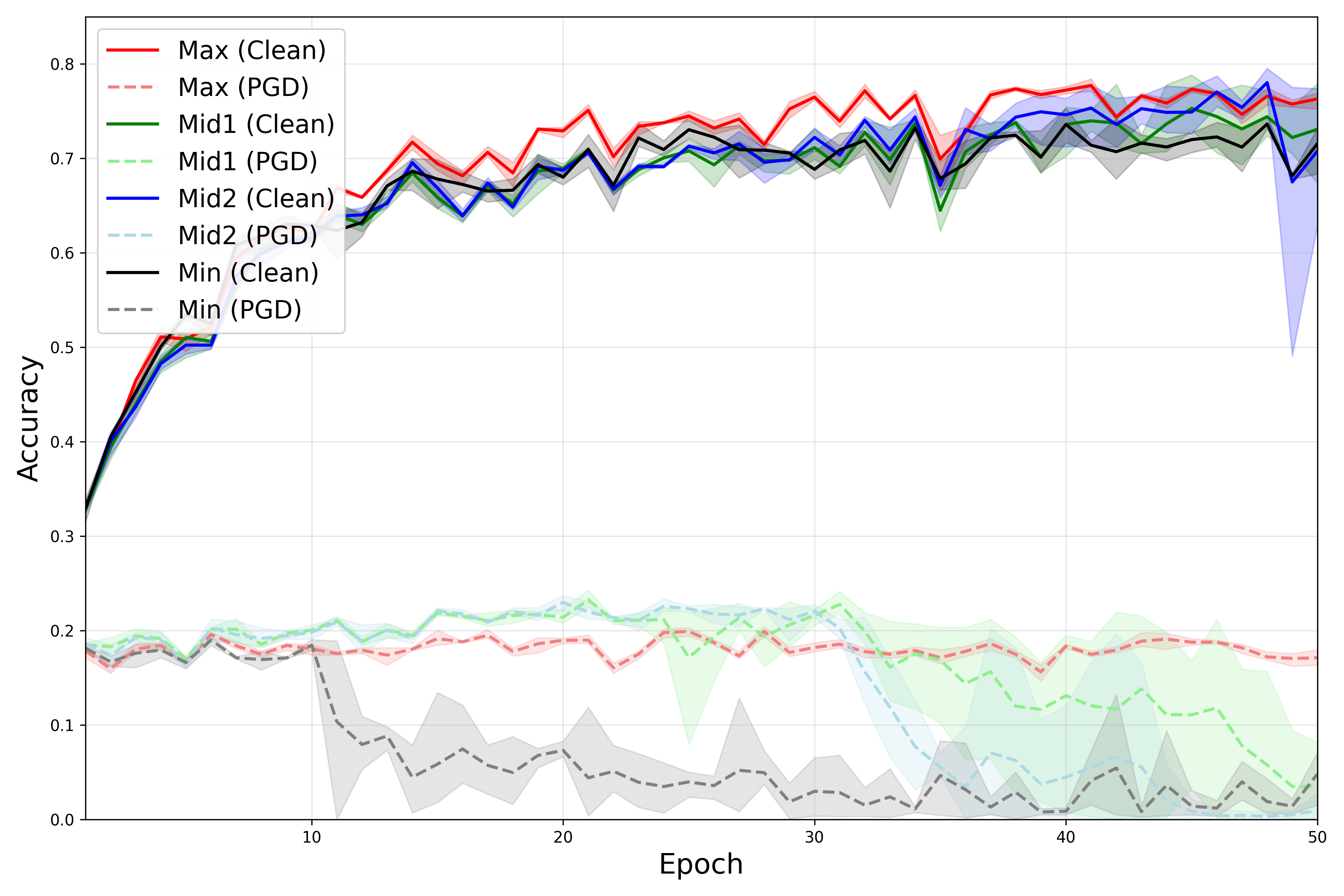}
    
    \vspace{-2mm}
    \caption{CO analysis on CIFAR10 and ResNet18. Each training batch is divided into four groups, Max, Min, Mid$_1$, and Mid$_2$. PGD use 10 steps, with a fixed perturbation budget of 8/255.}
    \label{co_analyze}
    \vspace{-2mm}
\end{figure}

In this section, we begin by analyzing the impact of guidance intensity on FAT performance.
Then, we describe the proposed distribution-aware dynamic guidance strategy.

\subsection{Impact of Guidance Intensity}
This section aims to answer the following question: \textit{Do different samples exhibit distinct behaviors during FAT, and if so, how?} To investigate this, we conduct a confidence-based analysis following the TDAT framework~\cite{tong2024taxonomy} on CIFAR-10 and ResNet-18. All experiments are performed with a training batch size of 128.

\begin{figure*}[htbp]
    \centering
    \begin{subfigure}[b]{0.95\textwidth}
        \centering
        \includegraphics[width=\textwidth,height=0.45\textheight,keepaspectratio]{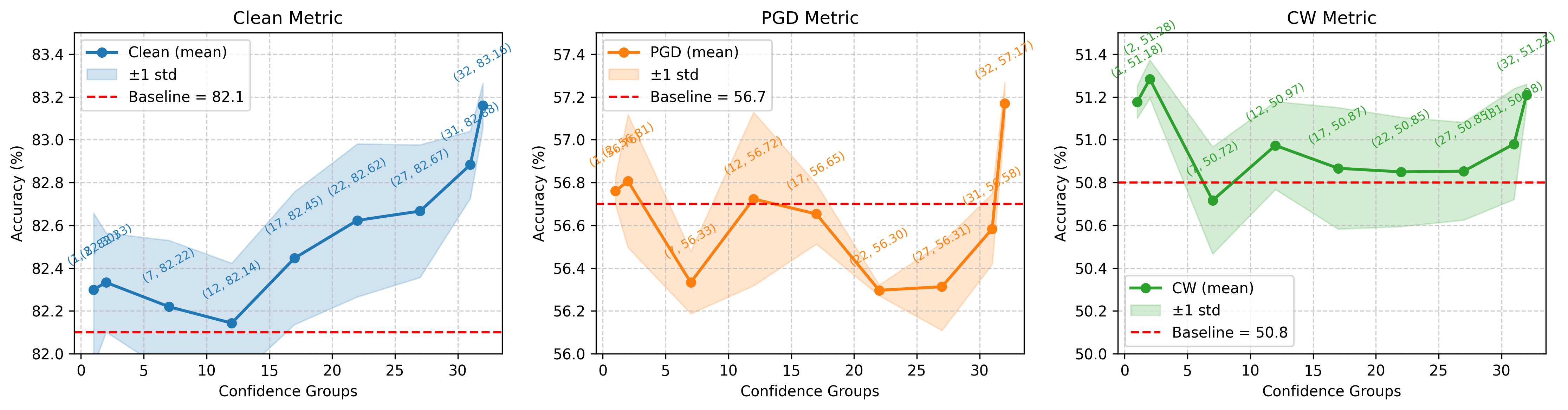}
        \captionsetup{position=top} 
        \caption{Setting the default $\xi$ to 8/255 and decreasing $\xi$ to 4/255 for the selected group (confidence descending order)}
    \end{subfigure}
    
    \vspace{0em}

    \begin{subfigure}[b]{0.95\textwidth}
        \centering
        \includegraphics[width=\textwidth,height=0.45\textheight,keepaspectratio]{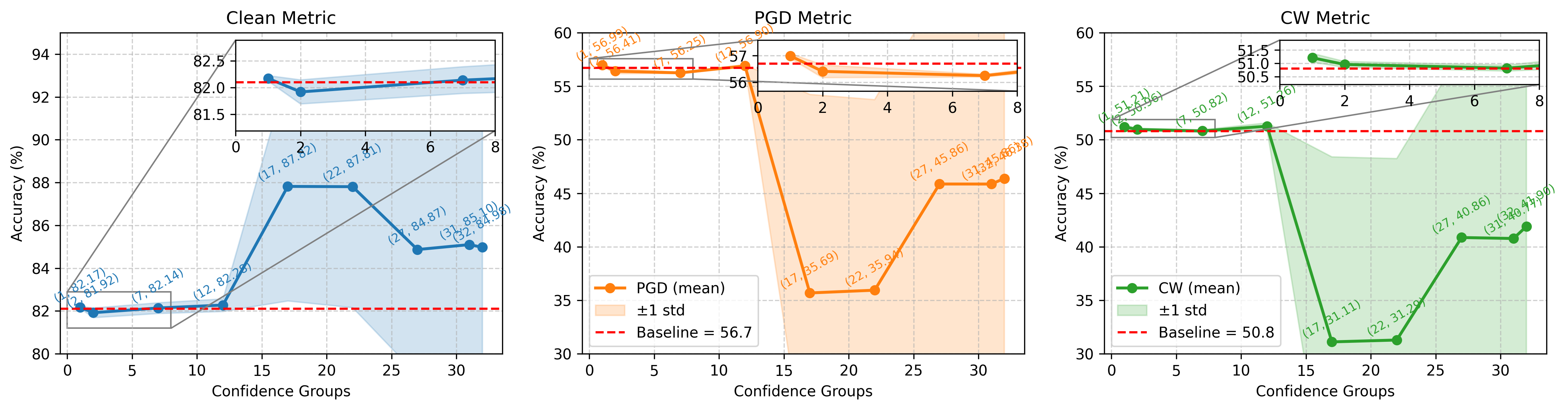}
        \captionsetup{position=top}
        \caption{Setting the default $\xi$ to 8/255 and increasing $\xi$ to 12/255 for the selected group (confidence descending order)}
    \end{subfigure}

    \caption{Trade-off analysis on CIFAR-10 with ResNet-18. PGD and C$\&$W use 10 steps, with a fixed perturbation budget of 8/255.}
    \label{tradeoff_analyze}
    \vspace{-2mm}
\end{figure*}

\subsubsection{Perturbation Strength Ablation}
\textit{CO Analysis.} 
To investigate how perturbation strength affects CO, batch samples are divided into four groups based on prediction confidence.
We selectively increase the perturbation budget to $16/255$ for the target group while maintaining $8/255$ for the remaining groups. The learning rate is fixed at 0.1.
Figure~\ref{co_analyze} presents two key findings:
1) High-confidence samples can tolerate large perturbations without destabilizing optimization; 
2) Low-confidence samples—predominantly misclassified—are more susceptible to CO when exposed to large perturbations.

\textit{Trade-off Analysis.} 
To further investigate the robustness–accuracy trade-off under varying perturbation strengths, each training batch is partitioned into 32 confidence groups—a finer granularity than that used in the CO analysis—to capture nuanced performance variations across confidence levels.
For the selected group, we adjust the perturbation budget—either decreasing it to $4/255$ or increasing it to $12/255$—while keeping other groups fixed at $8/255$. Figure~\ref{tradeoff_analyze} reveals four key observations:
1) Reducing the perturbation budget to $4/255$ consistently improves clean accuracy and moderately enhances robustness against C$\&$W;
2) Reducing the perturbation budget for the lowest-confidence group enhances both clean and robust accuracies, achieving gains of +1.06, +0.47, and +0.41 on Clean, PGD, and C$\&$W metrics, respectively.
This suggests that alleviating excessive error reinforcement on already misclassified samples can enhance overall performance;
3) Consistent with the CO analysis, applying large perturbations to low-confidence samples can trigger CO, indicating that even a small fraction of unstable samples may destabilize training;
4) Increasing the perturbation budget for high-confidence groups slightly improves model robustness, albeit sometimes at the expense of clean accuracy.


\begin{figure*}[htbp]
    \centering
    \begin{subfigure}[b]{0.95\textwidth}
        \centering
        \includegraphics[width=\textwidth,height=0.45\textheight,keepaspectratio]{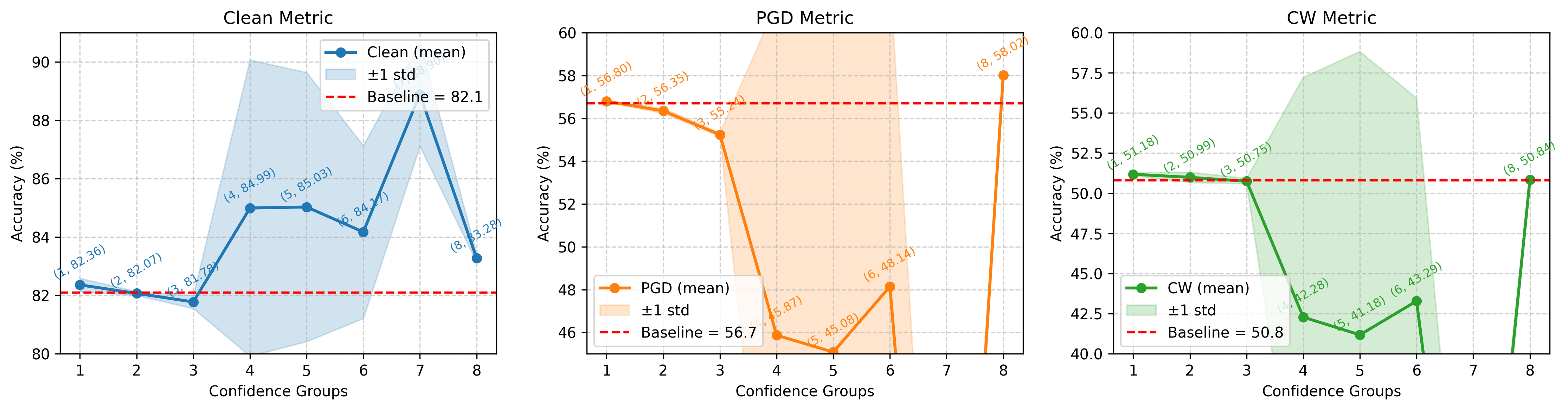}
        \captionsetup{position=top} 
        \caption{Setting $\beta_1$ to 0 and $\beta_2$ to -0.1 (confidence descending order)}
    \end{subfigure}
    
    \vspace{0em}

    \begin{subfigure}[b]{0.95\textwidth}
        \centering
        \includegraphics[width=\textwidth,height=0.45\textheight,keepaspectratio]{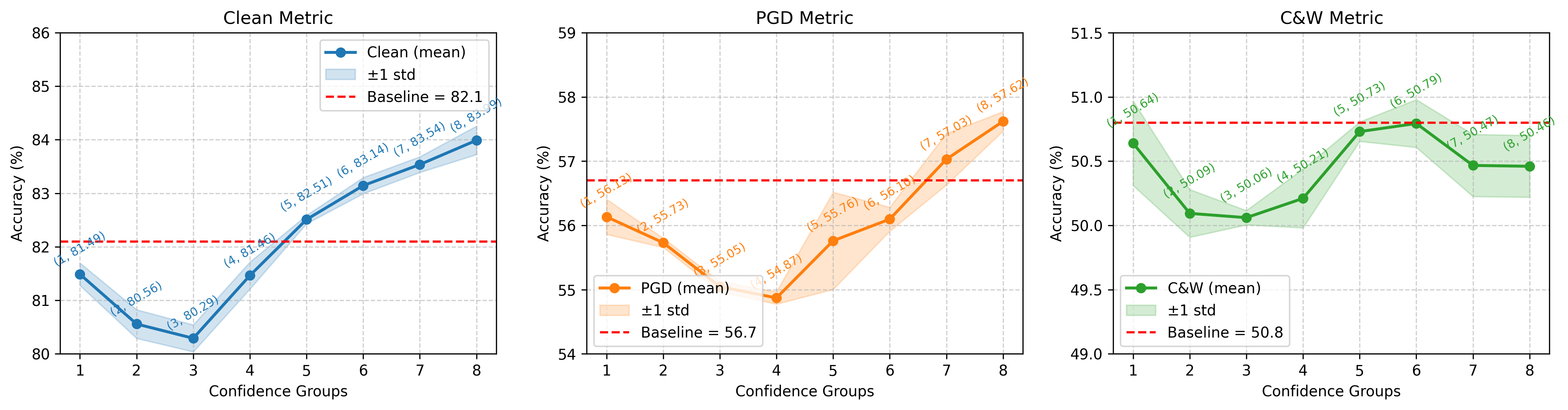}
        \captionsetup{position=top}
        \caption{Setting $\beta_1$ to 0.1 and $\beta_2$ to -0.1 (confidence descending order)}
    \end{subfigure}
    \vspace{-2mm}
    \caption{Supervision strength analysis on CIFAR-10 with ResNet-18. PGD and C$\&$W use 10 steps, with a perturbation budget of 8/255.}
    \label{supervision_analyze}
    \vspace{-2mm}
\end{figure*}

\subsubsection{Supervision Strength Ablation}
Recent studies have introduced label relaxation to improve adversarial training stability, typically formulated as
\begin{equation}\label{eq3}
    \mathbf{\hat{y}} = \mathbf{y} \cdot \gamma + \mathbf{1}\cdot\frac{1 - \gamma}{L},
\end{equation}
where $\mathbf{\hat{y}}$ denotes the relaxation label, $\mathbf{y}$ is the one-hot ground-truth label, $L$ signifies the number of classes, and $\gamma$ controls the relaxation strength.
Given that datasets vary in class cardinality, our analysis primarily focuses on the true class and the most probable incorrect class. Specifically, each training batch is divided into 8 confidence groups, and for the selected group, its supervision label is modified as
\begin{equation}\label{eq41}
    \mathbf{{y}^\prime} = \mathbf{\hat{y}} + \mathbf{y} \cdot \beta_1 +  \mathbf{y_m} \cdot \beta_2,
\end{equation}
where $\mathbf{y_m}$ represents the one-hot vector of the most probable incorrect class, and $\beta_1, \beta_2$ modulate the emphasis on the true and most probable incorrect classes, respectively.

Figure \ref{supervision_analyze} reveals several key observations: (1)
Applying $\beta_1 = 0$ and $\beta_2 = -0.1$ to the lowest-confidence group notably enhances both clean and adversarial accuracies, yielding average absolute gains of +1.18 on clean accuracy and +1.40 on PGD robustness, while maintaining comparable C$\&$W robustness to the baseline.
This finding aligns with the perturbation-strength ablation, confirming that mitigating excessive error reinforcement for already misclassified samples can improve model performance;
(2) When the same setting is applied to mid-confidence samples, training becomes unstable—likely because the membership of this confidence group changes rapidly during training, leading to inconsistent supervision signals and oscillatory gradient directions across iterations;
(3) Increasing $\beta_1$ improves the training stability, implying that reinforcing correct class supervision stabilizes FAT;
(4) Overly large $\beta_1$ degrades performance, as it induces overconfident predictions and amplifies regularization penalties.

\textbf{Remarks.} The above analyses indicate that applying uniform guidance to all samples is inherently suboptimal. Instead, both perturbation and supervision strengths should be adaptively adjusted based on each sample’s characteristics, guiding inputs with varying confidence levels toward more consistent decision boundaries. \textit{In particular, mitigating excessive error reinforcement for low-confidence (mostly misclassified) samples yields the most substantial improvements in both robustness and clean accuracy.}

\subsection{Distribution-aware Dynamic Guidance}
DDG jointly adjusts the perturbation budget and supervisory guidance, enabling adaptive alignment of decision boundaries while mitigating the over-reinforcement of erroneous signals.
Unlike prior methods that utilize a fixed perturbation budget ($\xi_\text{base} $= 8/255), DDG allocates budgets based on the confidence ranking $r_i$ of each example $x_i$:
\begin{equation}\label{eq4}
\xi_i = \xi_{\text{base}} + \kappa \left[ \tanh\left(r_i - \tau_1\right) - \tanh\left(\tau_2 - r_i\right) \right],
\end{equation}
where $\kappa$ serves as a scaling factor that controls the amplitude of dynamic variation (defaulting to $2/255$, which yields $\xi_{i} \in [4/255, 12/255]$). $\tau_1$ and $\tau_2$ govern transition regions of perturbation intensity, constrained by $\tau_1 + \tau_2 = B$, where $B$ represents the batch size. 
$r_i$ indicates the sample ranking, sorted in descending order of prediction confidence.
Eq.~(\ref{eq4}) assigns larger budgets to high-confidence samples and smaller budgets to low-confidence ones.
The adversarial perturbations are then clipped using $\xi_{i}$ as
\begin{equation}\label{eq5}
\delta_i = \operatorname{clip}\Big(
\delta_\text{init} + \max\{\xi_{i}, \xi_\text{base}\}\cdot\operatorname{sign}(\nabla_{x_i+\delta_\text{init}} \mathcal{L}),\
-\xi_{i},\ \xi_{i}
\Big).
\end{equation}
Here, \(\delta_\text{init}\) denotes the initial perturbation, typically inherited from the previous optimization step. The term \(\nabla_{x_i+\delta_\text{init}} \mathcal{L}\) is the gradient of the loss function with respect to the input \(x_i+\delta_\text{init}\). The operation \(\max\{\xi_{i}, \xi_\text{base}\}\)  establishes a lower bound for the perturbation step size.
The final adversarial example is formulated as $x_i^\prime = \operatorname{clip}(x_i + \delta_i, 0, 1)$.


In addition, DDG redefines the soft supervision signal to maintain training stability and mitigate error reinforcement,
\begin{equation}\label{eq6}
\mathbf{y_{\text{sr}}} =
\begin{cases}
\mathbf{\hat{y}}, & \text{if } \arg\max f(x^\prime) = y, \\[6pt]
\mathbf{\hat{y}} + \gamma (1 - \text{Acc}) \mathbf{y}\, - \dfrac{1}{L} \mathbf{y_m}, & \text{otherwise,}
\end{cases}
\end{equation}
where \(\text{Acc}\) denotes the empirical accuracy of the current model on the training batch, providing a dynamic, global performance signal.
Term $\gamma (1 - \text{Acc}) \mathbf{y}$ performs positive-class reinforcement, decreasing with higher true-class confidence, while term $\dfrac{1}{L} \mathbf{y_m}$ implements negative-class suppression, inversely proportional to the number of classes.

With the established adversarial inputs $x^\prime$ and their corresponding supervision labels $\mathbf{y_{\text{sr}}}$, the optimization objective is defined as follows:
\begin{equation}\label{eq7}
\mathcal{L}_\text{total} = - \frac{1}{B} \sum \mathbf{y_{\text{sr}}} \log f(x^\prime) + \mathcal{L}_{\text{smo}},
\end{equation}
\begin{equation}\label{eq8}
\footnotesize
\mathcal{L}_{\text{smo}}
=
\big\|
f(x + \delta_\text{init}) - f(x')
\big\|_2
\left(
\lambda\frac{\max(\xi_{B}) - \xi_{B}}{\max(\xi_{B}) - \min(\xi_{B})}
+ \alpha\,y_{\text{false}}
+ 1
\right).
\end{equation}
where $\|\cdot\|_2$ indicates the $\ell_2$-norm distance metric, $\max(\xi_{B})$ and $\min(\xi_{B})$ correspond to the maximum and minimum perturbation budgets within the current batch respectively, $\xi_{B}$ refers to the dynamic perturbation budget for batch samples, and $y_{\text{false}}$ serves as the misclassification indicator that returns 1 when $\arg\max f(x') \neq y$ and 0 otherwise.
The smoothing regularization term \(\mathcal{L}_{\text{smo}}\) comprises three complementary components: the term\(\lambda\frac{\max(\xi_{B}) - \xi_{B}}{\max(\xi_{B}) - \min(\xi_{B})}\) balances the regularization strength across the batch; the misclassification penalty $\alpha\, y_{\text{false}}$ reinforces regularization on misclassified examples, maintaining gradient smoothness throughout the negative-suppression process; and a constant term restricts non-zero regularization for all samples.

The Algorithm details of DDG are given in Algorithm \ref{alg:ddg}.


\begin{algorithm}[t]
\caption{The proposed DDG}
\label{alg:ddg}
\small
\SetAlgoLined
\KwIn{Training dataset $\mathcal{D}$, number of classes $L$, batch size $B$, model $f$ with parameters $\theta$, base perturbation budget $\xi_{\text{base}} = 8/255$, scaling factor $\kappa$, thresholds $\tau_1, \tau_2$, relaxation factor $\gamma$}
\KwOut{Trained model weights $\theta$}
Initialize model parameters $\theta$\;
\For{epoch = 1 to $N$}{
    \ForEach{batch $(X, y) \sim \mathcal{D}$}{
        Calculate confidence rankings of $X$ as $r$ \;

        \tcp{Assign Perturbation Budget:}
        
        $\xi = \xi_{\text{base}} + \kappa \left[ \tanh\left(r - \tau_1\right) - \tanh\left(\tau_2 - r\right) \right]$\;
        \tcp{Adv Example Generation:}
        $g = \nabla_{x+\delta_{\text{init}}} \mathcal{L}(f(x+\delta_{\text{init}}), y)$\;
            $\delta = \operatorname{clip}\left(\delta_{\text{init}} + \max\{\xi, \xi_{\text{base}}\} \cdot \operatorname{sign}(g), -\xi, \xi\right)$\;
            $x^\prime = \operatorname{clip}(x + \delta, 0, 1)$\;
        
\tcp{Assign Supervision Signal:}
        $\text{Acc} = \frac{1}{B}\sum\mathbb{I}[\arg\max f(x^\prime) = y]$\;
        $\mathbf{\hat{y}} = \mathbf{y} \cdot \gamma + \mathbf{1}\cdot\frac{1 - \gamma}{L}$\;
        
        $\mathbf{y_{\text{sr}}} =
\begin{cases}
\mathbf{\hat{y}}, & \text{if } \text{Pred True}, \\
\mathbf{\hat{y}} + \gamma (1 - \text{Acc}) \mathbf{y}\, - \dfrac{1}{L} \mathbf{y_m}, & \text{otherwise,}
\end{cases}$
        
        \tcp{Loss with Gradient Smoothness:}
        Compute regularization $\mathcal{L}_{\text{smo}}$ using Eq.~(\ref{eq8})\;
        $\mathcal{L}_{\text{total}} = -\frac{1}{B}\sum \mathbf{y_{\text{sr}}} \log f(x^\prime)$ +$\mathcal{L}_{\text{smo}}$ \;
        Update $\theta \leftarrow \theta - \eta \nabla_\theta \mathcal{L}_{\text{total}}$\;
    }
}
\Return{$\theta$}
\end{algorithm}
\section{Experiments}

\subsection{Experimental Settings}
\textit{General Setup.}
We evaluate our method on benchmark datasets: CIFAR-10~\cite{krizhevsky2009learning}, CIFAR-100~\cite{krizhevsky2009learning}, and Tiny ImageNet~\cite{deng2009imagenet}. 
To ensure fair comparison, all methods are trained under identical experimental settings following prior works~\cite{Wong2020, zhang2022revisiting, Jiag2022prior}. Specifically, we adopt ResNet-18~\cite{he2016deep} as the backbone for all datasets and train models using stochastic gradient descent (SGD) with a momentum of 0.9, weight decay of $5\times10^{-4}$, batch size of 128, and the initial learning rate of 0.1. Training is performed for 110 epochs, with the learning rate decayed by a factor of 0.1 at epochs 100 and 105. 
For each method, we report results at both the best and final epochs, where `best' denotes the epoch achieving the highest robustness against PGD-10, and `final' represents the last epoch to evaluate training stability.

\textit{Hyperparameters.}
The threshold \(\tau_1\) in Eq.~(\ref{eq4}) is set to 8. 
\(\lambda\) and \(\alpha\) in Eq.~(\ref{eq7}) are set to 1.33 and 1.5, respectively. Other hyperparameters follow the settings in prior work~\cite{tong2024taxonomy}.

\textit{Evaluation Protocol.}
Following the existing evaluation protocols~\cite{Wong2020, huang2023fast, tong2024taxonomy}, we utilize various attacks to evaluate adversarial robustness, including FGSM~\cite{goodfellow2014explaining}, BIM~\cite{kurakin2018adversarial}, PGD-10/20/50~\cite{madry2017towards}, C\&W~\cite{carlini2017towards}, APGD~\cite{Francesco2020}, and AutoAttack (AA)~\cite{Francesco2020}. Here, PGD-n refers to a projected gradient descent attack with $n$ iterations. AutoAttack combines APGD, FAB~\cite{5d1dce4b3a55ac56ce82a597}, and Square Attack~\cite{Andriushchenko20202}. All attacks are performed under the $\ell_\infty$ norm with a perturbation budget of $\xi=8/255$. For single-step attacks ($e.g.$ FGSM), the step size is 8/255; for multi-step attacks ($e.g.$ PGD), it is 2/255.

\textit{Baselines.}
We compare our work with recent FAT methods, including FGSM-RS~\cite{Wong2020}, FGSM-Free~\cite{shafahi2019adversarial}, FGSM-SDI~\cite{jia2022boosting}, GAT~\cite{sriramanan2020guided}, GradAlign~\cite{Andriushchenko2020}, N-FGSM~\cite{de2022make}, ZeroGrad~\cite{Golgooni2021}, LAS-AWP~\cite{jia2022adversarial}, NuAT~\cite{sriramanan2021towards}, ATAS~\cite{huang2023fast}, FGSM-PGI~\cite{Jiag2022prior}, FGSM-PGK~\cite{jia2024improving}, and TDAT~\cite{tong2024taxonomy}.

\begin{table*}[htpb]
\centering
\caption{Performance comparison of various AT methods on CIFAR-10. Bold numbers highlight the best results. 
`Steps': attack iterations during training. `Best': checkpoint with highest PGD-10 accuracy. `Final': last checkpoint.
}
\small
\label{tab1}
\setlength{\tabcolsep}{0.1cm}
\begin{tabular}{lccccccccccc}
\toprule
\multirow{2}{*}{Model} & \multirow{2}{*}{Steps} & \multirow{2}{*}{Type} & \multirow{2}{*}{Clean} & \multirow{2}{*}{FGSM~\cite{goodfellow2014explaining}} & \multirow{2}{*}{BIM~\cite{kurakin2018adversarial}} & \multicolumn{3}{c}{PGD~\cite{madry2017towards}} & \multirow{2}{*}{AA~\cite{Francesco2020}} & \multirow{2}{*}{C\&W~\cite{carlini2017towards}} & \multirow{2}{*}{APGD~\cite{Francesco2020}} \\
&&&&&& 10 & 20 & 50 \\
\midrule
\multirow{2}{*}{MART~\cite{wang2019improving}} & \multirow{2}{*}{10} & Best & 82.03 & 64.94 & 54.48 & 54.83 & 53.72 & 53.53 & 47.74 & 49.68 & 53.51  \\
 &  & Final & 82.33 & 65.12  & 53.90 & 54.38 & 52.98 & 52.60 & 47.46 & 49.66 & 52.77 \\
\multirow{2}{*}{LAS-AWP~\cite{jia2022adversarial}} & \multirow{2}{*}{10} & Best & 82.92 & 65.86  & 55.97 & 56.37 & 55.57 & 55.20 & 49.46 & 51.53 & 54.18 \\
 &  & Final & 82.92 & 65.86  & 55.97 & 56.37 & 55.57 & 55.20 & 49.46 & 51.53 & 54.18 \\
 \midrule
 
\multirow{2}{*}{FGSM-RS~\cite{Wong2020}} & \multirow{2}{*}{1} & Best & {\bf 83.69} & 62.00 & 47.20 & 47.66 & 46.29 & 45.96 & 42.80 & 46.10 & 46.13 \\
 &  & Final & 83.69 & 62.00 & 47.20 & 47.66 & 46.29 & 45.96 & 42.80 & 46.10 & 46.13 \\
\multirow{2}{*}{FGSM-Free~\cite{shafahi2019adversarial}} & \multirow{2}{*}{-} & Best & 81.38 & 60.81  & 48.74 & 49.07 & 48.03 & 47.62 & 44.37 & 46.98 & 47.90 \\
 &  & Final & 81.38 & 60.81 & 48.74 & 49.07 & 48.03 & 47.62 & 44.37 & 46.98 & 47.90 \\
\multirow{2}{*}{GAT~\cite{sriramanan2020guided}} & \multirow{2}{*}{1} & Best & 81.53 & 64.18 & 53.72 & 54.05 & 53.26 & 52.95 & 47.68 & 49.76 & 53.24 \\
 &  & Final & 81.88 & 64.30 & 52.89 & 53.23 & 52.16 & 51.86 & 47.08 & 49.71 & 52.05 \\
 
\multirow{2}{*}{FGSM-SDI~\cite{jia2022boosting}} & \multirow{2}{*}{1} & Best & 83.55 & 63.60 & 51.46 & 51.94 & 50.65 & 50.34 & 46.31 & 49.09 & 50.61 \\
 &  & Final & {\bf 83.73} & 63.75 & 51.28 & 51.88 & 50.49 & 50.09 & 46.34 & 49.42 & 50.43 \\
 
\multirow{2}{*}{GradAlign~\cite{Andriushchenko2020}} & \multirow{2}{*}{1} & Best & 80.45 & 60.56 & 48.80 & 49.11 & 47.96 & 47.63 & 43.92 & 46.94 & 47.85 \\
 &  & Final & 80.45 & 60.56 & 48.80 & 49.11 & 47.96 & 47.63 & 43.92 & 46.94 & 47.85 \\
 
\multirow{2}{*}{N-FGSM~\cite{de2022make}} & \multirow{2}{*}{1} & Best & 80.35 & 60.93 & 49.59 & 49.83 & 48.77 & 48.51 & 44.54 & 47.37 & 48.59 \\
 &  & Final & 80.35 & 60.93 & 49.59 & 49.83 & 48.77 & 48.51 & 44.54 & 47.37 & 48.59 \\
 
\multirow{2}{*}{FGSM-PGK~\cite{jia2024improving}} & \multirow{2}{*}{1} & Best & 81.52 & 64.95 & 55.28 & 56.14 & 55.58 & 55.36 & {\bf 48.94} & {\bf 50.90} & 55.44\\ 
 &  & Final & 81.63 & 65.01 & 55.02 & 55.79 & 55.34 & 55.09 & {\bf 48.85} & 50.75 & 55.28\\

 \multirow{2}{*}{FGSM-PGI~\cite{Jiag2022prior}} & \multirow{2}{*}{1} & Best & 81.71 & 65.02 & 54.87 & 55.26 & 54.54 & 54.38 & {48.60} & {50.88} & 54.62 \\
 &  & Final & 81.71 & 65.02 & 54.87 & 55.26 & 54.54 & 54.38 & { 48.60} & {\bf 50.88} & 54.62 \\

\multirow{2}{*}{TDAT~\cite{tong2024taxonomy}} & \multirow{2}{*}{1} & Best & 
82.46 & 66.28 & 55.87 & 56.36 & 55.63 & 55.41 & 48.06 & 49.99 & 55.10\\
 &  & Final &82.72 & 66.29 & 55.62 & 56.27 & 55.53 & 55.26 & 47.26 & 49.74 & 54.98 \\
\multirow{2}{*}{Ours} & \multirow{2}{*}{1} & Best & 82.67 & {\bf 68.44} & {\bf 59.79} & {\bf 60.44} & {\bf 59.57} & {\bf 59.33} & 48.08 & 49.86 & {\bf 59.48}\\
&&Final& 82.84 & {\bf 68.27} & {\bf 59.12} & {\bf 59.70} & {\bf 59.07} & {\bf 58.83} & 48.06 & 50.01 & {\bf 58.23}\\
\bottomrule
\end{tabular}
\vspace{-2mm}
\end{table*}

\subsection{Comparative Experiments and Analysis}

\textit{Results on CIFAR10.} 
Table~\ref{tab1} shows the experimental results on CIFAR10, with the observations given as follows:
1) DDG achieves the highest robustness across nearly all attack settings, reaching 68.44\%/59.48\%/59.33\% under FGSM/APGD/PGD-50 attacks—outperforming prior FAT methods such as TDAT (66.28\%/55.10\%/55.41\%).
2) DDG maintains consistent robustness between the best and final checkpoints (e.g., 60.44\% vs. 59.70\% on PGD-10), indicating effective mitigation of robust overfitting.
3) Despite using only single-step adversarial updates, DDG surpasses multi-step works such as MART (54.83\% on PGD-10) and LAS-AWP (56.37\% on PGD-10), while requiring only about one-tenth of their computational cost.
4) DDG shows a better trade-off between clean and robust accuracy. For instance, while FGSM-RS yields slightly higher clean accuracy (83.69\%), it exhibits notably lower robustness (47.66\% on PGD-10). Similarly, while FGSM-PGI and FGSM-PGK attain marginally higher robustness on C$\&$W and AA, they perform notably worse on other metrics.

\textit{Results on CIFAR-100.} Table~\ref{tab:robustness} reports the performance comparison of various FAT methods on CIFAR-100. Overall, our DDG achieves the strongest results among all competitors. Specifically, compared to recent advanced approaches TDAT (40.29\%/33.56\%/33.15\%) and FGSM-PGK (39.61\%/32.78\%/31.96\%), our method attains the highest robustness of 40.96\%/34.32\%/33.86\% against FGSM, PGD-10, and APGD attacks, respectively. Moreover, DDG achieves a clean accuracy of 57.98\%, surpassing TDAT (57.31\%) and FGSM-PGK (57.36\%), indicating that the robustness improvements do not compromise clean performance. These results validate the effectiveness of our work in balancing robustness–accuracy trade-off.

\textit{Results on Tiny-ImageNet.}
Table~\ref{tab:robustness_results} summarizes the performance of various FAT methods on Tiny-ImageNet. Our proposed approach achieves the optimal robustness across diverse adversarial attacks while maintaining competitive clean accuracy. Specifically, it attains the highest adversarial accuracy of 24.30\%/24.35\%/23.85\% under BIM, PGD-10, and APGD attacks, respectively, performing better than the strong baselines TDAT (23.78\%/23.98\%/23.28\%) and FGSM-PGI (23.19\%/23.27\%/22.91\%). In addition, our final model exhibits stable training behavior, yielding 23.40\%/23.64\%/22.95\% robustness under BIM, PGD-10, along with a clean accuracy of 44.83\%, which is comparable to or surpasses competing methods.

\begin{table*}[htpb]
\centering
\caption{Performance comparison of various AT methods on CIFAR-100. Bold numbers highlight the best results. 
}
\label{tab:robustness}
\setlength{\tabcolsep}{0.1cm}
\small
\begin{tabular}{lccccccccccc}
\toprule
\multirow{2}{*}{Methods} & \multirow{2}{*}{Steps} & \multirow{2}{*}{Type} & \multirow{2}{*}{Clean} & \multirow{2}{*}{FGSM~\cite{goodfellow2014explaining}} & \multirow{2}{*}{BIM~\cite{kurakin2018adversarial}} & \multicolumn{3}{c}{PGD~\cite{madry2017towards}} & \multirow{2}{*}{AA~\cite{Francesco2020}} & \multirow{2}{*}{C\&W~\cite{carlini2017towards}} & \multirow{2}{*}{APGD~\cite{Francesco2020}} \\
& & & & & & 10 & 20 & 50 \\
\midrule
\multirow{2}{*}{MART~\cite{wang2019improving}} & \multirow{2}{*}{10} & Best & 54.51 & 38.62 & 32.00 & 32.18 & 31.68 & 31.59 & 26.07 & 28.01 & 31.55 \\
&  & Final & 54.75 & 38.52  & 31.75 & 31.85 & 31.37 & 31.21 & 25.71 & 27.81 & 31.22 \\
\multirow{2}{*}{LAS-AWP~\cite{jia2022adversarial}} & \multirow{2}{*}{10} & Best & 58.75 & 40.66  & 32.42 & 32.58 & 31.91 & 31.74 & 27.23 & 29.59 & 31.74 \\
 &  & Final & 58.75 & 40.66 & 32.42 & 32.58 & 31.91 & 31.74 & 27.23 & 29.59 & 31.74 \\
 \midrule
\multirow{2}{*}{FGSM-RS~\cite{Wong2020}} & \multirow{2}{*}{1} & Best & 51.67 & 31.02 & 22.42 & 22.61 & 22.04 & 21.75 & 18.72 & 20.92 & 21.87 \\
 &  & Final & 51.67 & 31.02  & 22.42 & 22.61 & 22.04 & 21.75 & 18.72 & 20.92 & 21.87 \\
 
\multirow{2}{*}{FGSM-Free~\cite{shafahi2019adversarial}} & \multirow{2}{*}{-} & Best & 52.06 & 32.13 & 24.48 & 24.74 & 24.09 & 24.04 & 20.23 & 22.43 & 23.99  \\
 &  & Final & 52.06 & 32.13  & 24.48 & 24.74 & 24.09 & 24.04 & 20.23 & 22.43 & 23.99 \\
 
\multirow{2}{*}{GAT~\cite{sriramanan2020guided}} & \multirow{2}{*}{1} & Best & 57.49 & 36.77 & 28.91 & 29.14 & 28.60 & 28.30 & 23.11 & 25.14 & 28.42 \\
 &  & Final & 57.58 & 36.85 & 28.87 & 29.06 & 28.43 & 28.30 & 23.02 & 24.97 & 28.43 \\
 
\multirow{2}{*}{FGSM-SDI~\cite{jia2022boosting}} & \multirow{2}{*}{1} & Best & 58.64 & 37.23 & 28.60 & 28.78 & 27.99 & 27.67 & 23.27 & 25.85 & 27.83 \\
 &  & Final & 58.54 & 37.19  & 28.53 & 28.71 & 28.00 & 27.72 & 23.18 & 25.55 & 27.89  \\
\multirow{2}{*}{GradAlign~\cite{Andriushchenko2020}} & \multirow{2}{*}{1} & Best & 54.90 & 35.28  & 26.77 & 27.13 & 26.52 & 26.22 & 22.30 & 25.01 & 26.39 \\
 &  & Final & 55.22 & 35.51  & 26.82 & 27.12 & 26.42 & 26.24 & 22.19 & 24.94 & 26.52 \\
 
\multirow{2}{*}{N-FGSM~\cite{de2022make}} & \multirow{2}{*}{1} & Best & 54.41 & 35.00 & 26.99 & 27.01 & 26.55 & 26.34 & 22.81 & 25.08 & 26.31 \\
 &  & Final & 54.41 & 35.00  & 26.99 & 27.01 & 26.55 & 26.34 & 22.81 & 25.08 & 26.31 \\

 \multirow{2}{*}{FGSM-PGK~\cite{jia2024improving}} &\multirow{2}{*}{1} & Best & 57.36 & 39.61 & 32.12 & 32.78 & 32.35 & 32.19 & 25.84 & 28.40 & 31.96\\
& &Final& 57.36 & 39.61 & 32.12 & 32.78 & 32.35 & 32.19 & 25.84 & 28.40 & 31.96 \\

 \multirow{2}{*}{FGSM-PGI~\cite{Jiag2022prior}} & \multirow{2}{*}{1} & Best & {\bf 58.78} & 40.02 & 31.43 & 31.94 & 31.30 & 31.19 & 25.65 & 28.23 & 31.21\\
 &  & Final & {\bf 58.82} & 39.83 & 31.22 & 31.65 & 31.18 & 30.89 & 25.43 & 27.75 & 30.93 \\

\multirow{2}{*}{TDAT~\cite{tong2024taxonomy}} & \multirow{2}{*}{1} & Best & 57.32 & 40.29  & 33.33 & 33.56 & 33.17 & 33.06 & {\bf 26.61} & {\bf 28.47} & 33.15 \\
 &  & Final & 57.32 & 40.29 & 33.33 & 33.56 & 33.17 & 33.06 & 26.61 & {\bf 28.47} & 33.15 \\
 \multirow{2}{*}{Ours} & \multirow{2}{*}{1}&Best&57.98 & {\bf 40.96} & {\bf 34.11} & {\bf 34.42} & {\bf 33.92} & {\bf 33.79} & 26.54 & 28.42 & {\bf 33.86}\\
 &&Final & 58.15 & {\bf 41.17} & {\bf 33.70} & {\bf 33.99} & {\bf 33.63} & {\bf 33.51} & {\bf 26.79} & 28.46 & {\bf 33.48}\\
\bottomrule
\end{tabular}
\vspace{-2mm}
\end{table*}

\begin{table}[htpb]
\centering
\caption{Ablation analysis of components in DDG. PBA, SSA, and GS correspond to perturbation budget allocation, supervision signal adjustment, and gradient smoothness, respectively.}
\label{variant1}
\small
\setlength{\tabcolsep}{0.1cm}
\begin{tabular}{lcccccc}
\toprule
{Methods} &Type& {Clean} &FGSM& {PGD10} & {C\&W20} &APGD\\

\midrule
 \multirow{2}{*}{DDG}&Best & 82.67 &\textbf{68.44} & \textbf{60.44}&49.86&\textbf{59.48}\\
 &Avg&82.84&68.27&59.70&50.01&\textbf{58.23}\\
  \multirow{2}{*}{wo. PBA} &Best &81.70&68.33&59.54&49.58&57.46\\
 &Avg&81.56&68.23&\textbf{59.79}&49.54&57.79\\
 \multirow{2}{*}{wo. SSA}&Best &80.38&66.11&58.21&\textbf{50.87}&57.03\\
 &Avg&80.69&66.17&58.11&\textbf{50.63}&56.85\\
 \multirow{2}{*}{wo. GS} &Best& \textbf{84.32}&\textbf{68.44}&58.71&49.08&57.11\\
 &Avg&\textbf{84.55}&\textbf{68.34}&58.35&48.82&56.81\\
\bottomrule
\end{tabular}
\end{table}

\subsection{Ablation Studies}
\textit{Effects of each component.}
Table~\ref{variant1} presents the ablation results for the key components of DDG, including perturbation budget allocation (PBA), supervision signal adjustment (SSA), and gradient smoothness (GS). Removing PBA consistently degrades both clean and adversarial accuracies. Removing SSA improves robustness under the C$\&$W attack, but incurs a slight reduction in clean accuracy. In contrast, removing GS yields the highest clean accuracy (84.55\% on average), yet noticeably reduces adversarial performance. Overall, the complete DDG achieves the most favorable robustness–accuracy trade-off.

\begin{table}[htpb]
\centering
\caption{Ablation analysis of components in $\mathbf{y}_\text{sr}$. }
\label{variant2}
\small
\setlength{\tabcolsep}{0.1cm}
\begin{tabular}{lcccccc}
\toprule

\multirow{2}{*}{Scaling} & \multicolumn{3}{c}{Positive Adjustment} & \multicolumn{3}{c}{Negtive Adjustment}\\
&{Clean}& {PGD10} & {C\&W10}&{Clean}& {PGD10} & {C\&W10}\\
\midrule
0&82.03&59.85&50.47 &81.50&58.77&51.02\\
0.2&82.37&59.78&50.10 &81.39& 59.02& \textbf{51.16}\\
0.4&82.60&60.06&\textbf{50.71} &82.03 &59.36& 50.71\\
0.6&82.57&{59.96}&49.83&82.28 &59.44 &50.65\\
0.8&82.90&60.06&50.59 &82.39 &59.90 &50.48\\
1.0&83.04&\textbf{60.18}&50.64&82.94& 60.10 &50.75\\
1.2&\textbf{83.54}&60.16&50.29&\textbf{83.03}& \textbf{60.72}& 49.89\\
1.4&83.38&60.14&49.96&82.98& 60.65& 50.02\\
\bottomrule
\end{tabular}
\end{table}

\textit{Effects of components in $\mathbf{y}_\text{sr}$.}
In Eq.~(\ref{eq6}), $\mathbf{y}_\text{sr}$ contains both positive enhancement and negative suppression terms. To study their individual effects, we apply separate scaling factors (0 to 1.4, step 0.2), multiplying \(\gamma\) for the positive enhancement term and \(\frac{1}{L}\) for the negative suppression term. As reported in Table~\ref{variant2}, amplifying the negative suppression term consistently boosts clean accuracy and PGD robustness, while inducing only a slight drop in C$\&$W robustness. For instance, increasing the scaling factor from 0 to 1.0 yields gains of +1.44 in clean accuracy and +1.33  in PGD robustness, with  only a 0.27 reduction under the C$\&$W attack. Increasing the positive enhancement term exhibits a similar trend: clean accuracy improves steadily, whereas PGD and C$\&$W robustness change only marginally.

\begin{table*}[htpb]
\centering
\caption{Performance comparison of various AT methods on Tiny-ImageNet. Bold numbers highlight the best results. }
\label{tab:robustness_results}
\small
\setlength{\tabcolsep}{0.1cm}
\begin{tabular}{lccccccccccc}
\toprule
\multirow{2}{*}{Methods} & \multirow{2}{*}{Steps} & \multirow{2}{*}{Type} & \multirow{2}{*}{Clean} & \multirow{2}{*}{FGSM~\cite{goodfellow2014explaining}}  & \multirow{2}{*}{BIM~\cite{kurakin2018adversarial}} & \multicolumn{3}{c}{PGD~\cite{madry2017towards}} & \multirow{2}{*}{AA~\cite{Francesco2020}} & \multirow{2}{*}{C\&W~\cite{carlini2017towards}} & \multirow{2}{*}{APGD~\cite{Francesco2020}} \\
&&&&&& {10} & {20} & {50} \\
\midrule
\multirow{2}{*}{MART~\cite{wang2019improving}} & \multirow{2}{*}{10} & Best & 38.41 & 25.20 & 20.91 & 20.93 & 20.74 & 20.67 & 15.53 & 16.88 & 20.77 \\
 &  & Final & 36.83 & 18.02 & 12.25 & 12.36 & 12.01 & 11.93 & 9.26 & 10.36 & 12.02 \\
\multirow{2}{*}{LAS-AWP~\cite{jia2022adversarial}} & \multirow{2}{*}{10} & Best & 47.86 & 30.77 & 23.98 & 24.10 & 23.67 & 23.60 & 18.21 & 20.49 & 23.51\\
 &  & Final & 47.86 & 30.77 & 23.98 & 24.10 & 23.67 & 23.60 & 18.21 & 20.49 & 23.51 \\
 \midrule
\multirow{2}{*}{FGSM-RS~\cite{Wong2020}} & \multirow{2}{*}{1} & Best & 43.52 & 23.93 & 17.12 & 17.22 & 16.82 & 16.64 & 13.09 & 14.67 & 16.72\\
 &  & Final & 43.52 & 23.93 & 17.12 & 17.22 & 16.82 & 16.64 & 13.09 & 14.67 & 16.72 \\
\multirow{2}{*}{FGSM-Free~\cite{shafahi2019adversarial}} & \multirow{2}{*}{-} & Best & 44.15 & 25.18 & 17.81 & 17.95 & 17.47 & 17.31 & 13.67 & 15.82 & 17.32 \\
 &  & Final & 44.15 & 25.18 & 17.81 & 17.95 & 17.47 & 17.31 & 13.67 & 15.82 & 17.32 \\
\multirow{2}{*}{GAT~\cite{sriramanan2020guided}} & \multirow{2}{*}{1} & Best & 46.00 & 23.04 & 14.96 & 15.16 & 14.51 & 14.33 & 10.82 & 13.27 & 14.38 \\
 &  & Final & 45.57 & 22.10 & 14.38 & 14.56 & 14.03 & 13.85 & 10.26 & 12.71 & 13.90 \\
\multirow{2}{*}{FGSM-SDI~\cite{jia2022boosting}} & \multirow{2}{*}{1} & Best & 43.71 & 26.84 & 20.48 & 20.60 & 20.26 & 20.11 & 15.43 & 17.16 & 20.13  \\
 &  & Final & 45.40 & 24.76  & 17.15 & 17.27 & 16.84 & 16.73 & 12.47 & 14.80 & 16.97 \\

\multirow{2}{*}{GradAlign~\cite{Andriushchenko2020}} & \multirow{2}{*}{1} & Best & 38.22 & 22.85 & 17.13 & 17.20 & 16.86 & 16.79 & 12.64 & 14.00 & 16.85 \\
 &  & Final & 37.89 & 22.51  & 16.95 & 17.06 & 16.78 & 16.69 & 12.49 & 13.90 & 16.68 \\
 
\multirow{2}{*}{N-FGSM~\cite{de2022make}} & \multirow{2}{*}{1} & Best & {\bf 46.06} & 24.72  & 16.67 & 16.74 & 16.21 & 16.02 & 12.71 & 14.96 & 16.22 \\
 &  & Final & {\bf 46.06} & 24.72 & 16.67 & 16.74 & 16.21 & 16.02 & 12.71 & 14.96 & 16.22 \\
 \multirow{2}{*}{FGSM-PGI~\cite{Jiag2022prior}} & \multirow{2}{*}{1} & Best & 42.98 & 28.55 & 23.19 & 23.27 & 23.01 & 22.92 & {\bf 17.00} & 18.67 & 22.91 \\
 &  & Final & 45.13 & 28.11 & 21.43 & 21.51 & 21.19 & 21.07 & 14.86 & 16.86 & 21.14 \\
\multirow{2}{*}{TDAT~\cite{tong2024taxonomy}} & \multirow{2}{*}{1} & Best & 42.51 & {\bf 29.44} & 23.78 & 23.98 & 23.63 & 23.56 & 16.64 & 18.21 & 23.28\\
 &  & Final & 43.92 & 29.31 & 23.06 & 23.29& 22.85 & 22.74 & 15.87 & 17.54 & 22.50\\
 \multirow{2}{*}{Ours} & \multirow{2}{*}{1} & Best & 43.44 & 29.34 & {\bf 24.30} & {\bf 24.35} & {\bf 24.09} & {\bf 24.06} & 16.85 & {\bf 18.80} & {\bf 23.85}\\
  &  & Final & 44.83 & {\bf 29.74} & {\bf 23.40} & {\bf 23.64} & {\bf 23.24} & {\bf 23.07} & {\bf 15.94} & {\bf 17.85} & {\bf 22.95} \\
\bottomrule
\end{tabular}
\end{table*}

\begin{figure*}[t]
    \centering
        \includegraphics[width=0.98\textwidth]{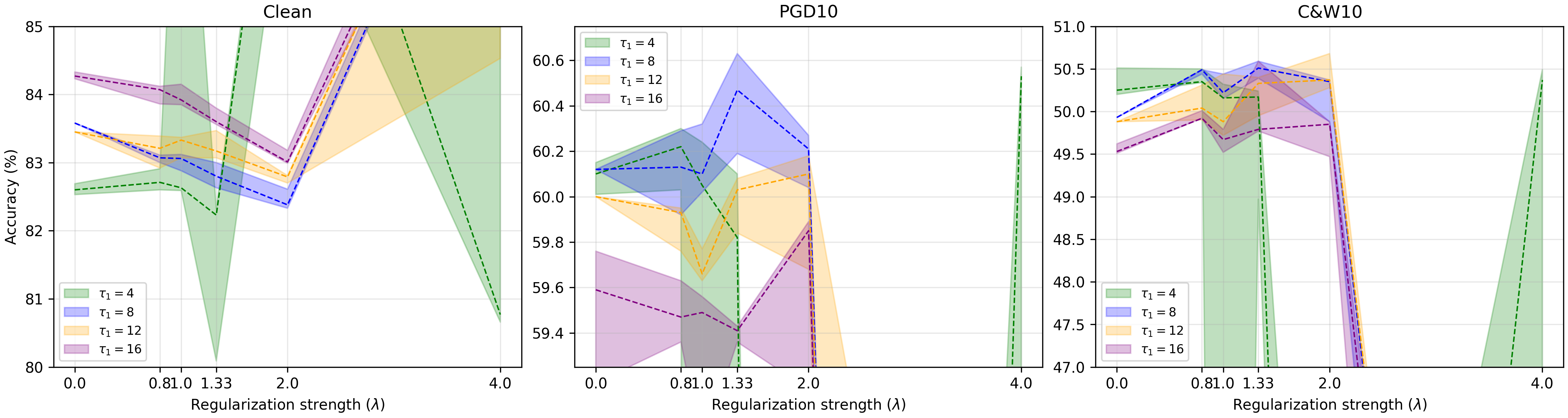}
    
    \caption{The impact of $\tau_1$ in Eq.~(\ref{eq4}) and $\lambda$ in Eq.~(\ref{eq7}) on FAT performance.}
    \label{analyze_tau_lambda}
    \vspace{-2mm}
\end{figure*}

Notably, clean accuracy is positively correlated with PGD robustness but negatively correlated with C$\&$W robustness. This discrepancy stems from the distinct attack objectives: PGD can be interpreted as a one-to-any attack that pushes samples away from the true class toward any incorrect class, whereas C$\&$W behaves as an any-to-one attack that identifies the easiest path. As clean accuracy increases, the model forms sharper class boundaries, making it more difficult to push a sample toward arbitrary incorrect classes—thereby improving PGD robustness. However, such boundaries may also reveal clearer descent directions toward specific target classes, leading to a slight reduction in C$\&$W robustness.

\textit{Impact of $\tau_1$ and $\lambda$.} Figure~\ref{analyze_tau_lambda} illustrates the effects of $\tau_1$ in Eq.~(\ref{eq4}) and $\lambda$ in Eq.~(\ref{eq7}) on FAT performance. Under stable training, clean accuracy increases as $\tau_1$ grows, while robust accuracy gradually decreases. Conversely, as $\lambda$ increases, the clean accuracy declines whereas the robust accuracy improves. Notably, the influence of these hyperparameters on overall performance is modest: clean, PGD, and C$\&$W accuracies vary within $83\%\pm0.5$, $60\%\pm0.5$, and $50.25\%\pm0.25$, respectively.





\section{Conclusions}
This paper introduces a Distribution-aware Dynamic Guidance (DDG) strategy to address two key challenges in fast adversarial training: catastrophic overfitting and the robustness–accuracy trade-off. Through systematic perturbation and supervision ablations, we show that applying uniform guidance across samples is suboptimal, leading to training instability and the learning of spurious correlations. To overcome these issues, DDG dynamically adjusts perturbation budgets according to sample confidence and modulates supervision signals based on each sample’s prediction state. Furthermore, an adaptive regularization term is introduced to maintain gradient smoothness under dynamic guidance. Experiments on multiple benchmarks demonstrate that DDG effectively resolves catastrophic overfitting and mitigates the robustness–accuracy trade-off.

\section*{Acknowledgements}
This work was supported by the National Natural Science Foundation of China under Grants 62431004, 62276046, and 62572426.

{
    \small
    \bibliographystyle{ieeenat_fullname}
    \bibliography{main}
}

\end{document}